\pdfoutput=1

\documentclass[11pt]{article}

\usepackage{acl}

\usepackage{times}
\usepackage{latexsym}

\usepackage{macros}

\usepackage[T1]{fontenc}

\usepackage[utf8]{inputenc}

\usepackage{microtype}

\usepackage{graphicx}
\graphicspath{{figures/}}
\usepackage{listings}
\usepackage{xcolor}
\usepackage{lipsum}
\definecolor{codegreen}{rgb}{0,0.6,0}
\definecolor{codegray}{rgb}{0.5,0.5,0.5}
\definecolor{codepurple}{rgb}{0.58,0,0.82}
\definecolor{backcolour}{rgb}{0.95,0.95,0.92}

\lstdefinestyle{mystyle}{
    backgroundcolor=\color{backcolour},   
    commentstyle=\color{codegreen},
    keywordstyle=\color{magenta},
    numberstyle=\tiny\color{codegray},
    numbers=none,
    stringstyle=\color{codepurple},
    basicstyle=\ttfamily\scriptsize,
    breakatwhitespace=false,         
    breaklines=false,                 
    captionpos=b,                    
    keepspaces=true,                 
    numbers=left,                    
    numbersep=5pt,                  
    showspaces=false,                
    showstringspaces=false,
    showtabs=false,                  
    tabsize=2
}
\title{PromptSource: An Integrated Development Environment \\
and Repository for Natural Language Prompts}

\author{
Stephen H. Bach$^{*1,2}$~~~%
Victor Sanh$^{*3}$~~~%
Zheng-Xin Yong$^1$~~~%
Albert Webson$^{1}$~~~%
Colin Raffel$^{3}$~~~%
\AND
Nihal V. Nayak$^1$~~~%
Abheesht Sharma$^4$~~~%
Taewoon Kim$^5$~~~%
M Saiful Bari$^6$~~~%
Thibault Fevry$^7$~~~%
\AND
Zaid Alyafeai$^8$~~~%
Manan Dey$^9$~~~%
Andrea Santilli$^{10}$~~~%
Zhiqing Sun$^{11}$~~~%
Srulik Ben-David$^{12}$~~~%
\AND
Canwen Xu$^{13}$~~~%
Gunjan Chhablani$^7$~~~%
Han Wang$^{14}$~~~%
Jason Alan Fries$^{15,2}$~~~%
\AND
Maged S. Al-shaibani$^{8}$~~~%
Shanya Sharma$^{16}$~~~%
Urmish Thakker$^{17}$~~~%
Khalid Almubarak$^{18}$~~~%
\AND
Xiangru Tang$^{19}$~~~%
Dragomir Radev$^{19}$~~~%
Mike Tian-Jian Jiang$^{20}$~~~%
Alexander M. Rush$^3$
\\
$^1$ Brown University~~~%
$^2$ Snorkel AI~~~%
$^3$ Hugging Face~~~%
$^4$ BITS Pilani~~~%
$^5$ VU Amsterdam~~~%
\\
$^6$ NTU~~~%
$^7$ BigScience~~~%
$^8$ KFUPM~~~%
$^9$ SAP~~~%
$^{10}$ University of Rome~~~%
$^{11}$ CMU~~~%
$^{12}$ Technion~~~%
\\
$^{13}$ UCSD~~~%
$^{14}$ NYU~~~%
$^{15}$ Stanford University~~~%
$^{16}$ Walmart Labs~~~%
$^{17}$ SambaNova Systems~~~%
\\
$^{18}$ PSAU~~~%
$^{19}$ Yale University~~~%
$^{20}$ ZEALS~~~%
$^*$ Equal Contribution~~~%
}

\begin{document}
\maketitle

\begin{abstract}
\textit{PromptSource} is a system for creating, sharing, and using natural language prompts.
Prompts are functions that map an example from a dataset to a natural language input and target output.
Using prompts to train and query language models is an emerging area in NLP that requires new tools that let users develop and refine these prompts collaboratively.
\textit{PromptSource} addresses the emergent challenges in this new setting with (1) a templating language for defining data-linked prompts, (2) an interface that lets users quickly iterate on prompt development by observing outputs of their prompts on many examples, and (3) a community-driven set of guidelines for contributing new prompts to a common pool. Over 2,000 prompts for roughly 170 datasets are already available in \textit{PromptSource}. \textit{PromptSource} is available at \url{https://github.com/bigscience-workshop/promptsource}.
\end{abstract}
\section{Introduction}

Prompt engineering is emerging as a new focus in NLP, particularly in zero- and few-shot learning settings.
\emph{Prompting} is the practice of representing a task as a natural language utterance in order to query a language model for a response~\citep{liu:corr2021}.
For example, if a language model is conditioned on the text \textit{``She hit a home run. The previous sentence is about ...''}, then the model's subsequent generation would be interpreted as a prediction of the topic of the preceding sentence, e.g. by mapping a response such as \textit{``sports''} to a class label.
In specific contexts, prompting has been shown to have advantages over traditional classification, for example facilitating adaptation of language models to ad-hoc tasks and improving sample efficiency in low-data settings~\citep{brown2020gpt3,schick-schutze-2021-just,le-scao-rush-2021-many,gao-etal-2021-making}. These advantages motivate a practical challenge: \emph{How can we enable users to create, refine, and share prompts?}

The process of prompt engineering is critical for successful deployment as choices in prompting can affect downstream predictions significantly, particularly in the zero-shot setting~\cite{perez:neuraips2021,zhao:arxiv2021,albert:arxiv2021}.
Furthermore, training directly on collections of prompts can enable large models to generalize to new prompts more robustly~\cite{sanh2021multitask,wei2021flan,metaicl,mishra:arxiv2021}.
There is therefore a growing need for tools that support the creation of corpora of prompts.

\textit{PromptSource} is an integrated development environment and repository for natural language prompts to use in the context of zero-shot (or gradient-based few-shot)  learning.
It provides a Web-based GUI that enables developers to write prompts in a templating language and immediately view their outputs on different examples. 
The system is integrated with the HuggingFace Datasets library~\citep{lhoest-etal-2021-datasets}, so that users can load any dataset automatically, browse existing prompts, and create new ones.
Through the course of writing thousands of prompts, we converged on three key aspects to the design of \textit{PromptSource}:
\begin{compactitem}
\item {\bf Flexible Templating Language.}
We adapt a templating language to represent prompts.
Prompt authors can define prompts in terms of dataset fields, hard-coded text, and simple control logic.
This choice provides the flexibility of a programming environment without the mental overhead of having to write and read arbitrary code. Prompt templates can easily be distributed and used in other systems.
\item {\bf Tools for Prompt Management.}
\textit{PromptSource} has multiple view to address the needs of prompt authors at different stages of the prompt engineering cycle.
A global view lets authors browse datasets and existing prompt templates.
A local view facilitates iteration on prompt wording and metadata, as well as testing on individual examples.

\item {\bf Community-Driven Quality Standards.}
 \textit{PromptSource} includes a set of guidelines for prompting based on a large-scale prompt writing pilot.
\textit{PromptSource}'s collection is meant to be useful for a wide range of research, based on iterative refinement of a set of quality standards. Prompts in \textit{PromptSource} are also annotated with various pieces of metadata to make finding and using prompts easier.
\end{compactitem}

The \textit{PromptSource} system includes over 2,000 open-source prompts for roughly 170 datasets, which have all been reviewed to meet the quality standards.
This collection, which we call the Public Pool of Prompts (P3), allows users to materialize prompted forms of datasets for hundreds of different tasks.
The T0 series of models~\citep{sanh2021multitask} for zero-shot inference were fine-tuned on a subset of P3.
Since then, \textit{PromptSource} and P3 have been extended for research on multi-lingual prompting~\citep{lin2021multilingual} and priming, i.e., in-context few-shot learning~\citep{metaicl}.
The \textit{PromptSource} system and associated content is a first step in the study of systems for prompt engineering, an area that is likely to continue to grow.

\section{Background and Related Work} \label{sec:related-work}

\textit{PromptSource} builds on recent work in prompting and prompt engineering.
It is also related to work on systems for other types of annotations.

\noindent {\bf Prompting~~}
Recently, prompting has emerged as a new focus within NLP as it can dramatically improve language models' few-shot and zero-shot performance in a wide range of downstream tasks~\citep{brown2020gpt3,timo:eacl2021,sanh2021multitask,wei2021flan}.
Prompts and prompt engineering come in several varieties~\citep{liu:corr2021}.
\textit{PromptSource} is focused on facilitating research with human-written prompts, in which natural language is the medium for describing tasks.
This approach has the advantage that prompts can be understood, modified, and applied without being tied to a specific model.
In contrast, past work has also aimed to automatically construct prompts by framing the search for a good prompt as a learning problem.
These prompts can either be expressed in natural language~\citep{gao-etal-2021-making,shin-etal-2020-autoprompt} or as arbitrary vectors (a.k.a. “continuous” or “soft” prompts) not corresponding to words in the model's original vocabulary~\citep{lester-etal-2021-power,qin-eisner-2021-learning}

When using human-written prompts, there are several possible approaches to learning.
One is a zero-shot setting, where the goal is to generalize to prompts for which no training examples are given.
Prompts can also be used in a few-shot setting, in which a model is either (1) trained on prompted examples of the target task via gradient updates, or (2) priming (i.e. in-context learning), in which labeled examples are included in an input sequence in order to prime models to make predictions without gradient updates~\citep{brown2020gpt3}. 

\textit{PromptSource} was originally designed for zero-shot learning, so it emphasizes explicit task instructions and no priming examples.
If needed, users can extend \textit{PromptSource} for few-shot learning (e.g., as done in \citealp{lin2021multilingual} and \citealp{metaicl}, described in \S\ref{sec:case-studies}).

\begin{figure*}[th!]
    \centering
    \includegraphics[width=1.0\textwidth]{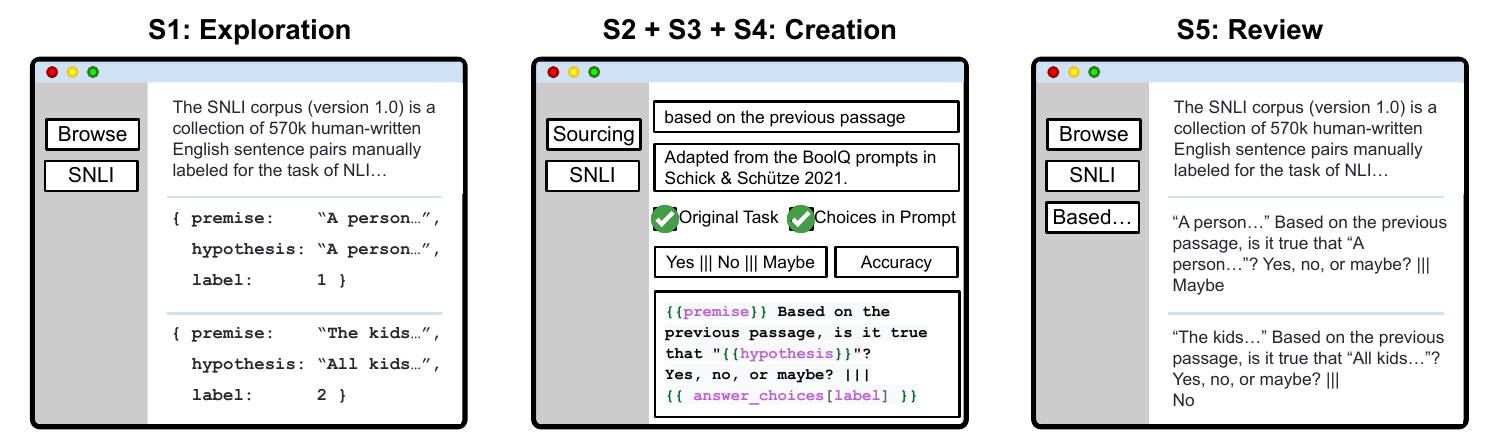}
    \caption{The five stages of creating prompts in \textit{PromptSource}.
     The Browse view for Dataset Exploration (S1).
    The Sourcing view for Prompt Writing (S2), Prompt Documentation (S3), and Iteration and Variation (S4).
    The Browse view for performing a Global Review (S5).}
    \label{fig:system}
\end{figure*}

\noindent {\bf Systems for Annotating Data~~}
Most work on collecting annotations has focused on labels and other annotations at the level of individual examples~\citep{neves:bioinformatics2021}.
GATE~\citep{cunningham2002gate} was an early system for annotating text, and includes support for many data types such as labels and entity tags.
Since then, many Web-based systems for annotating text have been developed~\citep{stenetorp:eacl12,saldago:bioinformatics12,wei:nucleicacidsresearch13,yimam:acl13,chen:naacl13,castilho:lt4dh16,putra:lrec20}.
Other systems support collaboration among multiple annotators~\citep{yang:acl18,stewart:emnlp19}.
More recently, many annotation systems have begun to incorporate learned models to improve workflow, using techniques such as active learning~\citep{lin:acl19,li:naacl21} and example recommendation~\citep{lee:acl20,kiela:naacl21}.
These systems are possible because the annotations to be collected are labels, for which metrics like inter-annotator agreement and model confidence are available.

There has also been some work on collecting annotations other than labels.
AlvisAE~\citep{papazian:acl12} and TreeAnnotator~\citep{helfrich:lrec18} support creating ontologies and other structured annotations.
Prompts differ from these annotations in that they are semi-structured functions, requiring new tools for developers.
\section{System Design and Workflow}

Creating prompts differs from other types of data collection and annotation. We focus on three challenging aspects on which prompting differs from traditional NLP annotation:

\begin{compactitem}
\item \textbf{Functions, not Labels.} A single prompt is a function that maps dataset examples (dictionaries of arbitrary fields) to natural language input/target pairs.
Creating a prompt is therefore more like programming than typical data annotation.
How should a prompt format trade off between expressivity and simplicity?
\item \textbf{Dataset-Level Choices.}
Prompts are associated with datasets, unlike label annotations that are local to single examples.
Prompt engineering requires developers to evaluate their choices across all examples.
What interfaces do authors need to inspect and debug their prompts?
\item \textbf{Variation in Prompt Construction.}
Unlike with labels, it is often desirable to have variation within prompt construction, 
as different prompt choices may lead to different results.
However, variation complicates quality judgment, and makes it impossible to apply simple metrics like inter-annotator agreement.
How can multiple authors collaborate to build a high-quality corpus of prompts and associated metadata?
\end{compactitem}

To illustrate these distinct aspects, we start with a concrete overview of the prompt creation process  of \textit{PromptSource}.
For this example, we imagine that a user of \textit{PromptSource} is creating prompts for a natural language inference dataset, specifically SNLI~\citep{bowman-etal-2015-large}.
The goal is to design a prompt query such that the answer can be mapped onto the SNLI classes. 
A prompt author can accomplish this goal with \textit{PromptSource} via the following five steps (Figure~\ref{fig:system}):

\textbf{S1: Dataset Exploration} The prompt author starts in the \emph{Browse} view to read the dataset description, including linked READMEs and papers, and to browse through examples. In this case, they would see that SNLI is a dataset for natural language inference: assume a given premise sentence is true, the goal is to determine whether a hypothesis sentence is true (entailment), false (contradiction), or undetermined (neutral). 

\textbf{S2: Prompt Writing} The prompt author uses the \emph{Sourcing} view to try out a prompt wording, and then adjusts it by observing prompted examples (Figure~\ref{fig:system} middle, full example in Figures~\ref{fig:sourcing} and~\ref{fig:sourcing_complex_jinja}).
 
\textbf{S3: Prompt Documentation} To facilitate using the prompt, the author fills in various metadata including possible metrics to evaluate the prompt, valid outputs if applicable, whether the prompt expresses the original intended task of the dataset, and whether the template explicitly states the valid outputs.

\textbf{S4: Iteration and Variation} The prompt author then iterates through S2 and S3 to create multiple prompts for the dataset. 
Authors are encouraged to vary multiple factors such as the formulation of the prompt and the targeted task (see Section~\ref{sec:contrib}).
 
\textbf{S5: Global Review} The author saves the draft prompts in a structured file which are then verified by other contributors through code reviews.
New prompts need to meet the quality standard with a series of automatic tests and by validation through prompted instances.
Upon passing review, the new prompts can be merged into a global prompts collection.

Upon submission, prompts can be viewed through \textit{PromptSource} by other users. The full collection is stored globally and can be used outside of the tool, for instance to be applied on an example from a dataset of the \textit{Datasets} library~\citep{lhoest-etal-2021-datasets}.

\begin{lstlisting}[language=python,numbers=none]
from promptsource.templates import DatasetTemplates
from datasets import load_dataset

prompts = DatasetTemplates("snli")
prompt_key = "based on the previous passage" 
p = prompts[prompt_key]

dataset = load_dataset("snli", split="train")
example = dataset[0]

result = p.apply(example)
print("INPUT: ", result[0])
print("TARGET: ", result[1])
\end{lstlisting}

\noindent With this workflow in mind, we next describe the key aspects of the \textit{PromptSource} system in greater detail. 
\section{Prompting Language}

A key design decision is the format for prompts.
Previous works on prompting tended to use code for specifying each prompt.
We experimented with this format and found a trade-off between expressivity and explicit structure. On one side, a maximally expressive format such as pure Python code would let users write complex programs to manipulate the semi-structured examples into prompted examples. However, interpreting and analyzing these programs becomes difficult. This difficulty limits downstream manipulation and analysis of the prompts, for example for possible future work on automatic prompt augmentation. On the other side, a maximally structured format, such as rule-based generation, limits the kinds of prompts that users can create. We found it infeasible to enumerate types of rules sufficient for the wide range of tasks and data formats for which we wanted prompts.

\begin{figure}[t]
    \centering
    \includegraphics[width=0.48\textwidth]{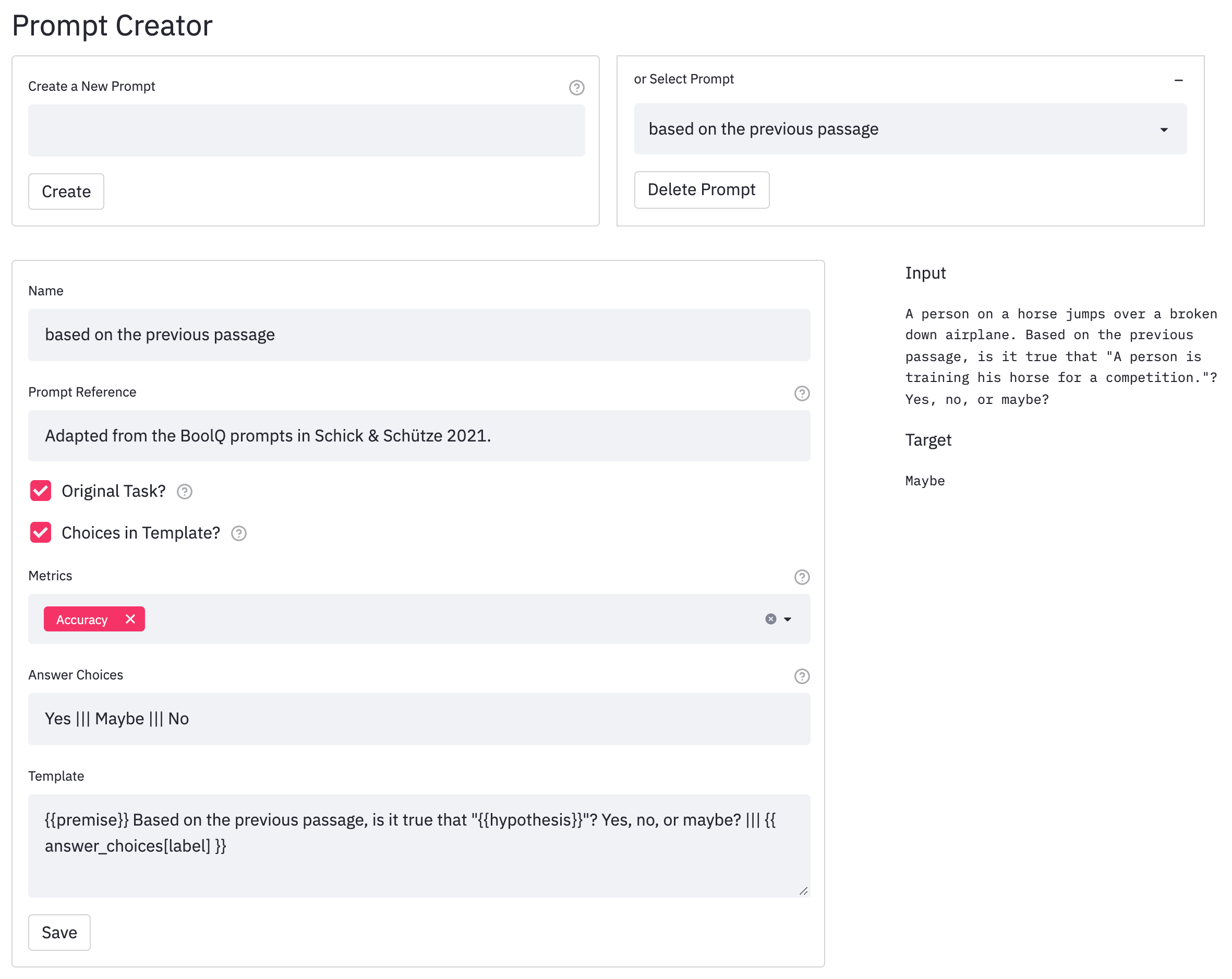}
    \caption{With the \textit{Sourcing} view, prompt authors can write new prompts, fill in the associated metadata, observe the result on examples, and iterate.}
    \label{fig:sourcing}
\end{figure}

We therefore settled on a middle ground between the two: a templating language. Specifically, we use the Jinja2 templating engine,\footnote{\url{https://jinja.palletsprojects.com}} originally designed for producing web markup. Users write templates as prompts with placeholders, such as \texttt{If \{\{premise\}\} is true, is it also true that \{\{hypothesis\}\}? ||| \{\{entailed\}\}}. The separator \texttt{|||} denotes the break between the conditioning text and the desired completion. Placeholders refer to fields in the underlying example (represented as a Python \texttt{dict} by \textit{Datasets} \cite{lhoest-etal-2021-datasets}). Users also have access to Jinja's built-in functions, such as manipulating strings and structured data. For each prompt, prompted examples are created by applying the prompt to all examples in the corresponding dataset. While Jinja is a complete programming language, our review guidelines encourage simple functions with minimal additional logic (see Figure~\ref{fig:sourcing} and~\ref{fig:sourcing_complex_jinja} for example).

During the development of \textit{PromptSource}, we found that a few idioms were particularly useful. First, not all templates are applicable to all examples in a dataset. Users can wrap templates in Jinja's built-in conditional statements, and any example that results in an empty prompted example is simply skipped. Second, many examples can be used to make multiple training instances, such as a question that has multiple valid answers. We therefore added a \texttt{choice} function that selects an element from a list in a way that can be controlled during dataset generation, such as picking a random element using a seeded random number generator or generating different prompts for each combination of elements in the template. Third, many tasks such as classification and binary question answering have a small set of possible valid completions, and it is common to make predictions for these tasks by scoring only the valid completions and returning the highest one~\citep{brown2020gpt3,sanh2021multitask,wei2021flan}. Users therefore can list the valid completions in a separate field and access them as a list in their prompts (displayed as \texttt{Answer choices} in Figure~\ref{fig:sourcing}). These completions are then explicitly available when evaluating predictions for these prompted examples.
\section{The \textit{PromptSource} UI}

The \textit{PromptSource} system is designed to enable prompt creators to view data (S1), write prompts in a standard format (S2, S3, and S4), and verify that their templates work correctly (S5). We implemented a lightweight interface for the tool in Streamlit\footnote{\url{https://streamlit.io/}} so that users could download, run locally in a web browser, and then upload their results to a central repository. Testing iterations of the interface on pilot template-writing tasks, we converged on three views for the interface. 

\begin{figure}[t]
    \centering
    \includegraphics[width=0.5\textwidth]{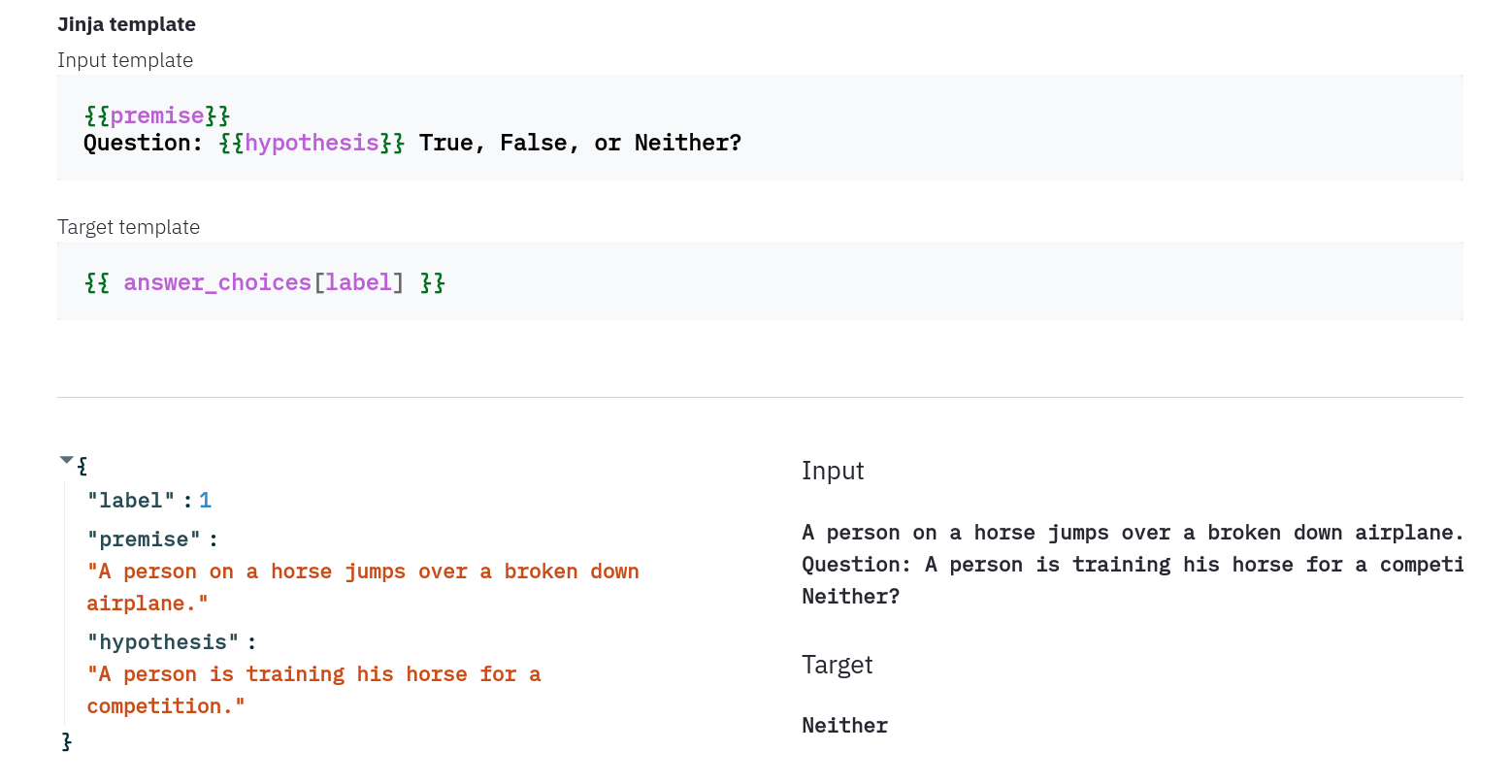}
    \caption{Prompt creators can browse through the dataset examples (left-column) and their prompted form (right column) using the \textit{Browse} view.}
    \label{fig:dataset_viewer}
\end{figure}

\textbf{V1: Browse} This view (Figure~\ref{fig:dataset_viewer}) lets users inspect datasets before creating prompts (S1).
Once prompts are created, they can select prompts and browse the prompted examples generated by them (S5). The original example is viewed side-by-side with the resulting prompted example, with the substituted text highlighted to distinguish from text hard-coded in the template. Users can quickly scroll through many examples, verify the behavior of their prompt, and return to the sourcing view if changes are needed.

\begin{figure}
    \centering
    \includegraphics[width=0.48\textwidth]{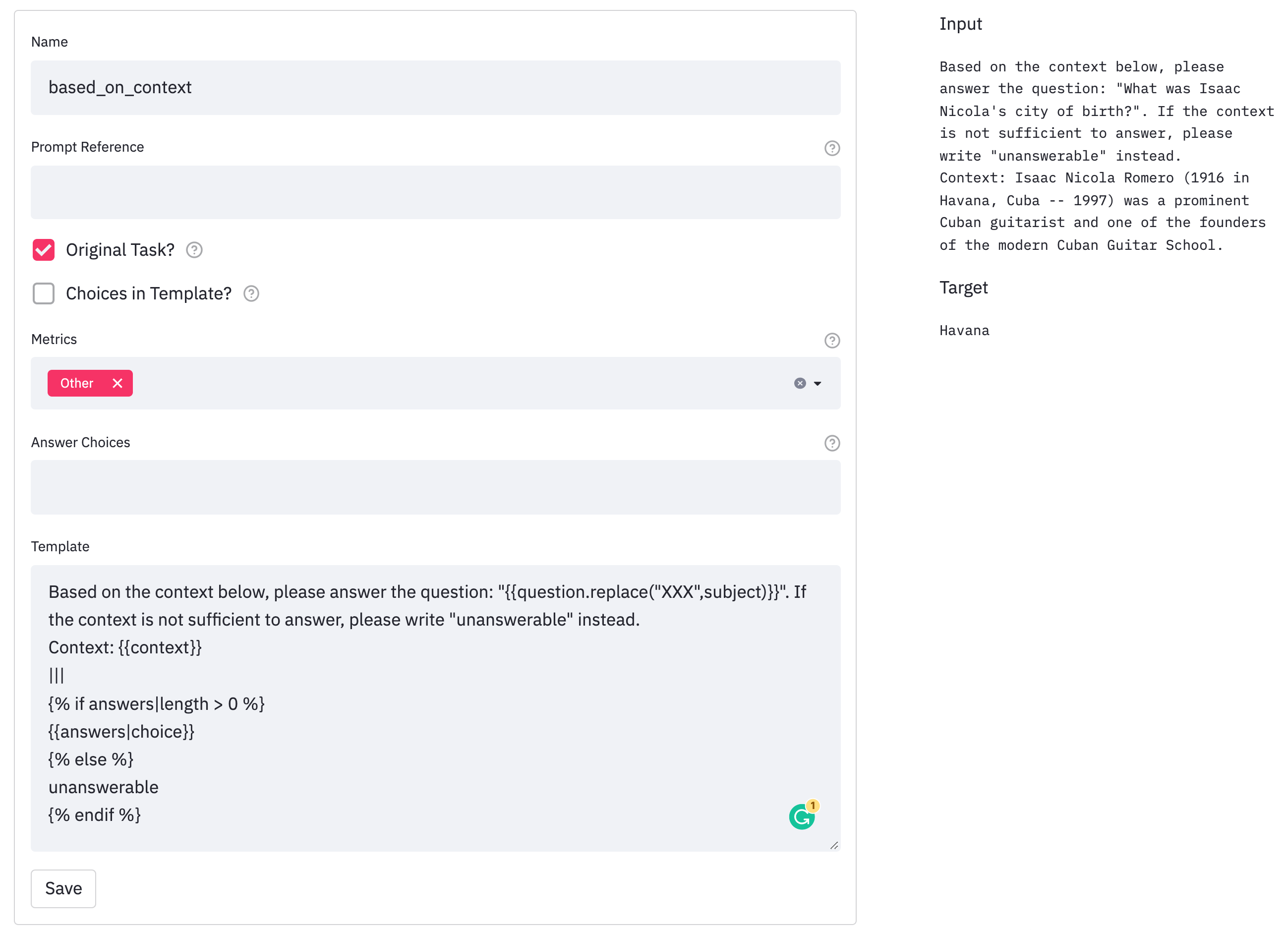}
    \caption{Another example of the the \textit{Sourcing} view, focusing on the editor.
    The templating language strikes a balance between expressivity and explicit structure. This prompt for QA-ZRE~\citep{levy-etal-2017-zero}, a dataset for zero-shot relation extraction, shows how to manipulate strings and do conditional statements with Jinja.}
    \label{fig:sourcing_complex_jinja}
\end{figure}

\textbf{V2: Sourcing} This view (Figures~\ref{fig:sourcing} and~\ref{fig:sourcing_complex_jinja}) allows users to select a dataset to prompt, browse examples from that dataset in the form of tables, and enter a prompt for that dataset. As the user writes their template (S2, S3, and S4), every time they save it, the output of the template applied to the current example is displayed next to the editor. We also collect metadata like a name for the template, and a reference for any bibliographic information or rationale for the template.

\begin{figure}[t]
    \centering
    \includegraphics[width=0.48\textwidth]{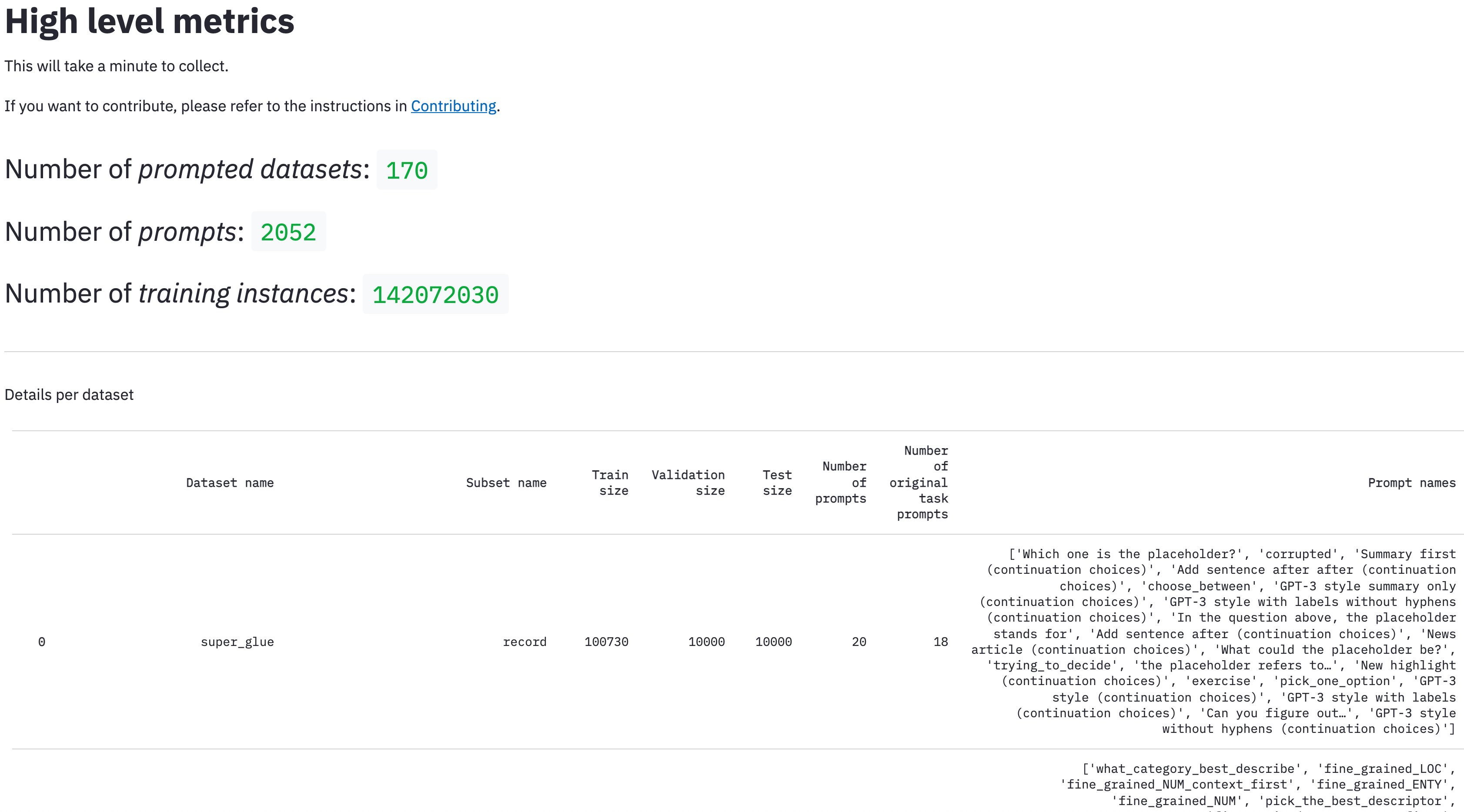}
    \caption{The \textit{Helicopter} view indicates what datasets have prompts and how many prompts are available for each dataset.}
    \label{fig:helicopter}
\end{figure}

\textbf{V3: Helicopter} This view (Figure~\ref{fig:helicopter}) allows users to see what datasets are available for writing templates and how many are written for each, to prioritize user attention. This view is particularly useful for moving between datasets and for the prompt reviewers (S5).

\section{Community Guidelines and Process}
\label{sec:contrib}

Due to the variety of existing NLP datasets, we found it challenging to exhaustively describe the characteristics of a good prompt: there are no simple metrics like inter-annotator agreement on example-level labels.
Instead, over a few iterations, we converged on community guidelines\footnote{Complete guidelines can be found at \url{https://github.com/bigscience-workshop/promptsource/blob/main/CONTRIBUTING.md}.} with three objectives in mind: (a) provide a standardized vocabulary for discussing prompts between prompt authors, reviewers and users, and minimum requirements for a valid prompt, (b) highlight common errors and best practices, (c) collect the necessary information about the prompts to support current and future research on prompt engineering. The guidelines were enforced in the use of \textit{PromptSource} by a code review process in which each prompt was reviewed before being committed to the central repository.

Guidelines apply to the combination of a template (a function that maps an example into an input/target pair in natural language) and a set of metadata about the template. The most important constraint we imposed for a template to be valid is that it is formulated in natural language (both for the input and the target). We forbid the use of non-natural language prompts such as pure code. Each prompt should clearly state what task should be solved, in a way a non-specialist adult can understand. We found this guideline strikes a good balance between freedom and expressivity in the wording of the prompts on one side and short generic prompts on the other side.

In early experiments, we found that user-written prompts that did not explicitly state the possible valid completions tended to perform worse in experiments than their counterparts in which the possible valid completions were listed. We encouraged prompt authors to explicitly state the valid outputs in some of their prompts.
In addition, when working with training prompts that include target text, we found it useful to remove variations on the target format that led to spurious ambiguity. For instance, the target template should only contain the answer to the task. It should not contain any extra text such as ``The answer is ...'', which can be equivalently moved to the input template.

One of the research question we hope to enable with \textit{PromptSource} is whether the diversity of the prompt formulation during training leads to models that are more robust to the prompt formulation at test time. Therefore, we encouraged prompt authors to create between 5 and 10 (or more) prompts per dataset while varying the prompt formulation. For a given dataset, authors produce multiple prompts per example, sometimes for task formulations that differed from the original dataset. For instance, for question answering dataset, one prompt can ask to extract the answer to a given question from a given passage, while a second prompt can ask to generate a potential question given an answer and a passage.

As part of the community process and to facilitate future research, \textit{PromptSource} asks prompt authors to include additional metadata for each prompt. Metadata fields include a name for the prompt, a reference to the paper it was extracted from (or any relevant explanation), whether the prompt expresses the task originally intended by the dataset, the valid outputs (if relevant), whether the input template states the valid outputs, and possible metrics to evaluate the prompted examples. These can be used in future systems to evaluate how the style and structure of prompts leads to different downstream results.

\section{Case Studies} \label{sec:case-studies}

A system for creating, maintaining, and using prompts is a key tool for supporting the emerging research area of prompting in a standardized and reproducible manner. We highlight three recent research projects for which \textit{PromptSource} was a key resource.

\noindent {\bf Massively multitask prompted training~~}
\citet{sanh2021multitask} study the question of zero-shot behaviors in large language models and ask whether zero-shot generalization can be induced by training a language model on a massively multitask mixture of prompts. To test this question, they use \textit{PromptSource} to create diverse prompts for a large collection of NLP datasets. Their training and evaluation prompts are a subset of P3.
This work demonstrates that \textit{PromptSource} allows training a language model on a massively multitask mixture of prompted datasets and evaluating the ability of models trained with such a procedure to perform unseen tasks.

\noindent {\bf Multilingual prompting~~}
\citet{lin2021multilingual} study the zero- and few-shot learning abilities of an multilingual autoregressive language model trained on 30 languages. In particular, they are interested in the cross-lingual generalization of such models and benchmark a variety of tasks in multiple languages. \textit{PromptSource} allows using a massive set of high-quality English prompts. Moreover, the English prompts serve as support to create prompts in other languages (through either machine or human translation).

\noindent {\bf Priming (in-context learning)~~}
\citet{metaicl} study improving models' few-shot priming performance by first fully training a model (with gradient updates) on a multitask mixture formatted with priming examples. They find that incorporating templates from P3 significantly further improves performance compared to training on priming examples alone. Although \textit{PromptSource} was not originally designed for this specific form of prompting, users were able to  easily use P3's template collection and the templating language for their own priming methods.

\section{Conclusion}

\textit{PromptSource} is an open-source system for creating, sharing, and using natural language prompts and addresses the need for new collaborative and centralized tools to support the emerging research around prompting. The tool is designed to answer three key needs: a flexible template language, a suite of tools for prompt management, and community-driven quality standards. As of January 2022, \textit{PromptSource} includes a growing collection of 2,000 public prompts for roughly 170 datasets, and has already been an instrumental resource for multiple recent research projects.

\section*{Acknowledgements}
This research was conducted under the BigScience project for open research,\footnote{\url{https://bigscience.huggingface.co/}} a year-long initiative targeting the study of large models and datasets.
The goal of the project is to research language models in a public environment outside large technology companies.
The project has over 950 researchers from over 65 countries and more than 250 institutions. The BigScience project was initiated by Thomas Wolf at Hugging Face, and this collaboration would not have been possible without his effort.
This research was the focus of the BigScience Prompt Engineering working group, which focused on the role of prompting in large language model training.
Disclosure: Stephen Bach contributed to this work as an advisor to Snorkel AI.

\bibliography{anthology,custom}

\begin{thebibliography}{37}
\expandafter\ifx\csname natexlab\endcsname\relax\def\natexlab#1{#1}\fi

\bibitem[{Bowman et~al.(2015)Bowman, Angeli, Potts, and
  Manning}]{bowman-etal-2015-large}
Samuel~R. Bowman, Gabor Angeli, Christopher Potts, and Christopher~D. Manning.
  2015.
\newblock \href {https://doi.org/10.18653/v1/D15-1075} {A large annotated
  corpus for learning natural language inference}.
\newblock In \emph{Proceedings of the 2015 Conference on Empirical Methods in
  Natural Language Processing}, pages 632--642, Lisbon, Portugal. Association
  for Computational Linguistics.

\bibitem[{Brown et~al.(2020)Brown, Mann, Ryder, Subbiah, Kaplan, Dhariwal,
  Neelakantan, Shyam, Sastry, Askell, Agarwal, Herbert{-}Voss, Krueger,
  Henighan, Child, Ramesh, Ziegler, Wu, Winter, Hesse, Chen, Sigler, Litwin,
  Gray, Chess, Clark, Berner, McCandlish, Radford, Sutskever, and
  Amodei}]{brown2020gpt3}
Tom~B. Brown, Benjamin Mann, Nick Ryder, Melanie Subbiah, Jared Kaplan,
  Prafulla Dhariwal, Arvind Neelakantan, Pranav Shyam, Girish Sastry, Amanda
  Askell, Sandhini Agarwal, Ariel Herbert{-}Voss, Gretchen Krueger, Tom
  Henighan, Rewon Child, Aditya Ramesh, Daniel~M. Ziegler, Jeffrey Wu, Clemens
  Winter, Christopher Hesse, Mark Chen, Eric Sigler, Mateusz Litwin, Scott
  Gray, Benjamin Chess, Jack Clark, Christopher Berner, Sam McCandlish, Alec
  Radford, Ilya Sutskever, and Dario Amodei. 2020.
\newblock \href
  {https://proceedings.neurips.cc/paper/2020/hash/1457c0d6bfcb4967418bfb8ac142f64a-Abstract.html}
  {Language models are few-shot learners}.
\newblock In \emph{Advances in Neural Information Processing Systems 33: Annual
  Conference on Neural Information Processing Systems 2020, NeurIPS 2020,
  December 6-12, 2020, virtual}.

\bibitem[{Chen and Styler(2013)}]{chen:naacl13}
Wei-Te Chen and Will Styler. 2013.
\newblock \href {https://aclanthology.org/N13-3004} {{A}nafora: A web-based
  general purpose annotation tool}.
\newblock In \emph{Proceedings of the 2013 {NAACL} {HLT} Demonstration
  Session}, pages 14--19, Atlanta, Georgia. Association for Computational
  Linguistics.

\bibitem[{Cunningham et~al.(2002)Cunningham, Maynard, Bontcheva, and
  Tablan}]{cunningham2002gate}
Hamish Cunningham, Diana Maynard, Kalina Bontcheva, and Valentin Tablan. 2002.
\newblock \href {https://doi.org/10.3115/1073083.1073112} {{GATE}: an
  architecture for development of robust {HLT} applications}.
\newblock In \emph{Proceedings of the 40th Annual Meeting of the Association
  for Computational Linguistics}, pages 168--175, Philadelphia, Pennsylvania,
  USA. Association for Computational Linguistics.

\bibitem[{Eckart~de Castilho et~al.(2016)Eckart~de Castilho,
  M{\'u}jdricza-Maydt, Yimam, Hartmann, Gurevych, Frank, and
  Biemann}]{castilho:lt4dh16}
Richard Eckart~de Castilho, {\'E}va M{\'u}jdricza-Maydt, Seid~Muhie Yimam,
  Silvana Hartmann, Iryna Gurevych, Anette Frank, and Chris Biemann. 2016.
\newblock \href {https://aclanthology.org/W16-4011} {A web-based tool for the
  integrated annotation of semantic and syntactic structures}.
\newblock In \emph{Proceedings of the Workshop on Language Technology Resources
  and Tools for Digital Humanities ({LT}4{DH})}, pages 76--84, Osaka, Japan.
  The COLING 2016 Organizing Committee.

\bibitem[{Gao et~al.(2021)Gao, Fisch, and Chen}]{gao-etal-2021-making}
Tianyu Gao, Adam Fisch, and Danqi Chen. 2021.
\newblock \href {https://doi.org/10.18653/v1/2021.acl-long.295} {Making
  pre-trained language models better few-shot learners}.
\newblock In \emph{Proceedings of the 59th Annual Meeting of the Association
  for Computational Linguistics and the 11th International Joint Conference on
  Natural Language Processing (Volume 1: Long Papers)}, pages 3816--3830,
  Online. Association for Computational Linguistics.

\bibitem[{Helfrich et~al.(2018)Helfrich, Rieb, Abrami, L{\"u}cking, and
  Mehler}]{helfrich:lrec18}
Philipp Helfrich, Elias Rieb, Giuseppe Abrami, Andy L{\"u}cking, and Alexander
  Mehler. 2018.
\newblock \href {https://aclanthology.org/L18-1308} {{T}ree{A}nnotator:
  Versatile visual annotation of hierarchical text relations}.
\newblock In \emph{Proceedings of the Eleventh International Conference on
  Language Resources and Evaluation ({LREC} 2018)}, Miyazaki, Japan. European
  Language Resources Association (ELRA).

\bibitem[{Kiela et~al.(2021)Kiela, Bartolo, Nie, Kaushik, Geiger, Wu, Vidgen,
  Prasad, Singh, Ringshia, Ma, Thrush, Riedel, Waseem, Stenetorp, Jia, Bansal,
  Potts, and Williams}]{kiela:naacl21}
Douwe Kiela, Max Bartolo, Yixin Nie, Divyansh Kaushik, Atticus Geiger,
  Zhengxuan Wu, Bertie Vidgen, Grusha Prasad, Amanpreet Singh, Pratik Ringshia,
  Zhiyi Ma, Tristan Thrush, Sebastian Riedel, Zeerak Waseem, Pontus Stenetorp,
  Robin Jia, Mohit Bansal, Christopher Potts, and Adina Williams. 2021.
\newblock \href {https://doi.org/10.18653/v1/2021.naacl-main.324} {Dynabench:
  Rethinking benchmarking in {NLP}}.
\newblock In \emph{Proceedings of the 2021 Conference of the North American
  Chapter of the Association for Computational Linguistics: Human Language
  Technologies}, pages 4110--4124, Online. Association for Computational
  Linguistics.

\bibitem[{Le~Scao and Rush(2021)}]{le-scao-rush-2021-many}
Teven Le~Scao and Alexander Rush. 2021.
\newblock \href {https://doi.org/10.18653/v1/2021.naacl-main.208} {How many
  data points is a prompt worth?}
\newblock In \emph{Proceedings of the 2021 Conference of the North American
  Chapter of the Association for Computational Linguistics: Human Language
  Technologies}, pages 2627--2636, Online. Association for Computational
  Linguistics.

\bibitem[{Lee et~al.(2020)Lee, Khanna, Lin, Lee, Ye, Boschee, Neves, and
  Ren}]{lee:acl20}
Dong-Ho Lee, Rahul Khanna, Bill~Yuchen Lin, Seyeon Lee, Qinyuan Ye, Elizabeth
  Boschee, Leonardo Neves, and Xiang Ren. 2020.
\newblock \href {https://doi.org/10.18653/v1/2020.acl-demos.42} {{LEAN}-{LIFE}:
  A label-efficient annotation framework towards learning from explanation}.
\newblock In \emph{Proceedings of the 58th Annual Meeting of the Association
  for Computational Linguistics: System Demonstrations}, pages 372--379,
  Online. Association for Computational Linguistics.

\bibitem[{Lester et~al.(2021)Lester, Al-Rfou, and
  Constant}]{lester-etal-2021-power}
Brian Lester, Rami Al-Rfou, and Noah Constant. 2021.
\newblock \href {https://doi.org/10.18653/v1/2021.emnlp-main.243} {The power of
  scale for parameter-efficient prompt tuning}.
\newblock In \emph{Proceedings of the 2021 Conference on Empirical Methods in
  Natural Language Processing}, pages 3045--3059, Online and Punta Cana,
  Dominican Republic. Association for Computational Linguistics.

\bibitem[{Levy et~al.(2017)Levy, Seo, Choi, and
  Zettlemoyer}]{levy-etal-2017-zero}
Omer Levy, Minjoon Seo, Eunsol Choi, and Luke Zettlemoyer. 2017.
\newblock \href {https://doi.org/10.18653/v1/K17-1034} {Zero-shot relation
  extraction via reading comprehension}.
\newblock In \emph{Proceedings of the 21st Conference on Computational Natural
  Language Learning ({C}o{NLL} 2017)}, pages 333--342, Vancouver, Canada.
  Association for Computational Linguistics.

\bibitem[{Lhoest et~al.(2021)Lhoest, Villanova~del Moral, Jernite, Thakur, von
  Platen, Patil, Chaumond, Drame, Plu, Tunstall, Davison, {\v{S}}a{\v{s}}ko,
  Chhablani, Malik, Brandeis, Le~Scao, Sanh, Xu, Patry, McMillan-Major, Schmid,
  Gugger, Delangue, Matussi{\`e}re, Debut, Bekman, Cistac, Goehringer, Mustar,
  Lagunas, Rush, and Wolf}]{lhoest-etal-2021-datasets}
Quentin Lhoest, Albert Villanova~del Moral, Yacine Jernite, Abhishek Thakur,
  Patrick von Platen, Suraj Patil, Julien Chaumond, Mariama Drame, Julien Plu,
  Lewis Tunstall, Joe Davison, Mario {\v{S}}a{\v{s}}ko, Gunjan Chhablani,
  Bhavitvya Malik, Simon Brandeis, Teven Le~Scao, Victor Sanh, Canwen Xu,
  Nicolas Patry, Angelina McMillan-Major, Philipp Schmid, Sylvain Gugger,
  Cl{\'e}ment Delangue, Th{\'e}o Matussi{\`e}re, Lysandre Debut, Stas Bekman,
  Pierric Cistac, Thibault Goehringer, Victor Mustar, Fran{\c{c}}ois Lagunas,
  Alexander Rush, and Thomas Wolf. 2021.
\newblock \href {https://doi.org/10.18653/v1/2021.emnlp-demo.21} {Datasets: A
  community library for natural language processing}.
\newblock In \emph{Proceedings of the 2021 Conference on Empirical Methods in
  Natural Language Processing: System Demonstrations}, pages 175--184, Online
  and Punta Cana, Dominican Republic. Association for Computational
  Linguistics.

\bibitem[{Li et~al.(2021)Li, Yu, Quangang, and Liu}]{li:naacl21}
Yanzeng Li, Bowen Yu, Li~Quangang, and Tingwen Liu. 2021.
\newblock \href {https://doi.org/10.18653/v1/2021.naacl-demos.5}
  {{FITA}nnotator: A flexible and intelligent text annotation system}.
\newblock In \emph{Proceedings of the 2021 Conference of the North American
  Chapter of the Association for Computational Linguistics: Human Language
  Technologies: Demonstrations}, pages 35--41, Online. Association for
  Computational Linguistics.

\bibitem[{Lin et~al.(2019)Lin, Lee, Xu, Lan, and Ren}]{lin:acl19}
Bill~Yuchen Lin, Dong-Ho Lee, Frank~F. Xu, Ouyu Lan, and Xiang Ren. 2019.
\newblock \href {https://doi.org/10.18653/v1/P19-3010} {{A}lpaca{T}ag: An
  active learning-based crowd annotation framework for sequence tagging}.
\newblock In \emph{Proceedings of the 57th Annual Meeting of the Association
  for Computational Linguistics: System Demonstrations}, pages 58--63,
  Florence, Italy. Association for Computational Linguistics.

\bibitem[{Lin et~al.(2021)Lin, Mihaylov, Artetxe, Wang, Chen, Simig, Ott,
  Goyal, Bhosale, Du, Pasunuru, Shleifer, Koura, Chaudhary, O'Horo, Wang,
  Zettlemoyer, Kozareva, Diab, Stoyanov, and Li}]{lin2021multilingual}
Xi~Victoria Lin, Todor Mihaylov, Mikel Artetxe, Tianlu Wang, Shuohui Chen,
  Daniel Simig, Myle Ott, Naman Goyal, Shruti Bhosale, Jingfei Du, Ramakanth
  Pasunuru, Sam Shleifer, Punit~Singh Koura, Vishrav Chaudhary, Brian O'Horo,
  Jeff Wang, Luke Zettlemoyer, Zornitsa Kozareva, Mona~T. Diab, Veselin
  Stoyanov, and Xian Li. 2021.
\newblock \href {http://arxiv.org/abs/2112.10668} {Few-shot learning with
  multilingual language models}.
\newblock \emph{CoRR}, abs/2112.10668.

\bibitem[{Liu et~al.(2021)Liu, Yuan, Fu, Jiang, Hayashi, and
  Neubig}]{liu:corr2021}
Pengfei Liu, Weizhe Yuan, Jinlan Fu, Zhengbao Jiang, Hiroaki Hayashi, and
  Graham Neubig. 2021.
\newblock \href {https://arxiv.org/abs/2107.13586} {Pre-train, prompt, and
  predict: A systematic survey of prompting methods in natural language
  processing}.
\newblock \emph{CoRR}, abs/2107.13586.

\bibitem[{Min et~al.(2021)Min, Lewis, Zettlemoyer, and Hajishirzi}]{metaicl}
Sewon Min, Mike Lewis, Luke Zettlemoyer, and Hannaneh Hajishirzi. 2021.
\newblock \href {http://arxiv.org/abs/2110.15943} {Metaicl: Learning to learn
  in context}.
\newblock \emph{CoRR}, abs/2110.15943.

\bibitem[{Mishra et~al.(2021)Mishra, Khashabi, Baral, and
  Hajishirzi}]{mishra:arxiv2021}
Swaroop Mishra, Daniel Khashabi, Chitta Baral, and Hannaneh Hajishirzi. 2021.
\newblock Cross-task generalization via natural language crowdsourcing
  instructions.
\newblock \emph{arXiv preprint arXiv:2104.08773}.

\bibitem[{Neves and {\v{S}}eva(2021)}]{neves:bioinformatics2021}
Mariana Neves and Jurica {\v{S}}eva. 2021.
\newblock An extensive review of tools for manual annotation of documents.
\newblock \emph{Briefings in bioinformatics}, 22(1):146--163.

\bibitem[{Papazian et~al.(2012)Papazian, Bossy, and
  N{\'e}dellec}]{papazian:acl12}
Fr{\'e}d{\'e}ric Papazian, Robert Bossy, and Claire N{\'e}dellec. 2012.
\newblock \href {https://aclanthology.org/W12-3621} {{A}lvis{AE}: a
  collaborative web text annotation editor for knowledge acquisition}.
\newblock In \emph{Proceedings of the Sixth Linguistic Annotation Workshop},
  pages 149--152, Jeju, Republic of Korea. Association for Computational
  Linguistics.

\bibitem[{Perez et~al.(2021)Perez, Kiela, and Cho}]{perez:neuraips2021}
Ethan Perez, Douwe Kiela, and Kyunghyun Cho. 2021.
\newblock \href {https://arxiv.org/abs/2105.11447} {True few-shot learning with
  language models}.
\newblock \emph{NeurIPS}.

\bibitem[{Putra et~al.(2020)Putra, Teufel, Matsumura, and
  Tokunaga}]{putra:lrec20}
Jan Wira~Gotama Putra, Simone Teufel, Kana Matsumura, and Takenobu Tokunaga.
  2020.
\newblock \href {https://aclanthology.org/2020.lrec-1.854} {{TIARA}: A tool for
  annotating discourse relations and sentence reordering}.
\newblock In \emph{Proceedings of the 12th Language Resources and Evaluation
  Conference}, pages 6912--6920, Marseille, France. European Language Resources
  Association.

\bibitem[{Qin and Eisner(2021)}]{qin-eisner-2021-learning}
Guanghui Qin and Jason Eisner. 2021.
\newblock \href {https://doi.org/10.18653/v1/2021.naacl-main.410} {Learning how
  to ask: Querying {LM}s with mixtures of soft prompts}.
\newblock In \emph{Proceedings of the 2021 Conference of the North American
  Chapter of the Association for Computational Linguistics: Human Language
  Technologies}, pages 5203--5212, Online. Association for Computational
  Linguistics.

\bibitem[{Salgado et~al.(2012)Salgado, Krallinger, Depaule, Drula, Tendulkar,
  Leitner, Valencia, and Marcelle}]{saldago:bioinformatics12}
David Salgado, Martin Krallinger, Marc Depaule, Elodie Drula, Ashish~V.
  Tendulkar, Florian Leitner, Alfonso Valencia, and Christophe Marcelle. 2012.
\newblock \href {https://doi.org/10.1093/bioinformatics/bts435} {{MyMiner: a
  web application for computer-assisted biocuration and text annotation}}.
\newblock \emph{Bioinformatics}, 28(17):2285--2287.

\bibitem[{Sanh et~al.(2021)Sanh, Webson, Raffel, Bach, Sutawika, Alyafeai,
  Chaffin, Stiegler, Scao, Raja, Dey, Bari, Xu, Thakker, Sharma, Szczechla,
  Kim, Chhablani, Nayak, Datta, Chang, Jiang, Wang, Manica, Shen, Yong, Pandey,
  Bawden, Wang, Neeraj, Rozen, Sharma, Santilli, Fevry, Fries, Teehan,
  Biderman, Gao, Bers, Wolf, and Rush}]{sanh2021multitask}
Victor Sanh, Albert Webson, Colin Raffel, Stephen~H. Bach, Lintang Sutawika,
  Zaid Alyafeai, Antoine Chaffin, Arnaud Stiegler, Teven~Le Scao, Arun Raja,
  Manan Dey, M~Saiful Bari, Canwen Xu, Urmish Thakker, Shanya~Sharma Sharma,
  Eliza Szczechla, Taewoon Kim, Gunjan Chhablani, Nihal Nayak, Debajyoti Datta,
  Jonathan Chang, Mike Tian-Jian Jiang, Han Wang, Matteo Manica, Sheng Shen,
  Zheng~Xin Yong, Harshit Pandey, Rachel Bawden, Thomas Wang, Trishala Neeraj,
  Jos Rozen, Abheesht Sharma, Andrea Santilli, Thibault Fevry, Jason~Alan
  Fries, Ryan Teehan, Stella Biderman, Leo Gao, Tali Bers, Thomas Wolf, and
  Alexander~M. Rush. 2021.
\newblock \href {http://arxiv.org/abs/2110.08207} {Multitask prompted training
  enables zero-shot task generalization}.

\bibitem[{Schick and Sch{\"u}tze(2021{\natexlab{a}})}]{timo:eacl2021}
Timo Schick and Hinrich Sch{\"u}tze. 2021{\natexlab{a}}.
\newblock \href {https://doi.org/10.18653/v1/2021.eacl-main.20} {Exploiting
  cloze-questions for few-shot text classification and natural language
  inference}.
\newblock In \emph{Proceedings of the 16th Conference of the European Chapter
  of the Association for Computational Linguistics: Main Volume}, pages
  255--269, Online. Association for Computational Linguistics.

\bibitem[{Schick and
  Sch{\"u}tze(2021{\natexlab{b}})}]{schick-schutze-2021-just}
Timo Schick and Hinrich Sch{\"u}tze. 2021{\natexlab{b}}.
\newblock \href {https://doi.org/10.18653/v1/2021.naacl-main.185} {It{'}s not
  just size that matters: Small language models are also few-shot learners}.
\newblock In \emph{Proceedings of the 2021 Conference of the North American
  Chapter of the Association for Computational Linguistics: Human Language
  Technologies}, pages 2339--2352, Online. Association for Computational
  Linguistics.

\bibitem[{Shin et~al.(2020)Shin, Razeghi, Logan~IV, Wallace, and
  Singh}]{shin-etal-2020-autoprompt}
Taylor Shin, Yasaman Razeghi, Robert~L. Logan~IV, Eric Wallace, and Sameer
  Singh. 2020.
\newblock \href {https://doi.org/10.18653/v1/2020.emnlp-main.346}
  {{A}uto{P}rompt: {E}liciting {K}nowledge from {L}anguage {M}odels with
  {A}utomatically {G}enerated {P}rompts}.
\newblock In \emph{Proceedings of the 2020 Conference on Empirical Methods in
  Natural Language Processing (EMNLP)}, pages 4222--4235, Online. Association
  for Computational Linguistics.

\bibitem[{Stenetorp et~al.(2012)Stenetorp, Pyysalo, Topi{\'c}, Ohta, Ananiadou,
  and Tsujii}]{stenetorp:eacl12}
Pontus Stenetorp, Sampo Pyysalo, Goran Topi{\'c}, Tomoko Ohta, Sophia
  Ananiadou, and Jun{'}ichi Tsujii. 2012.
\newblock \href {https://aclanthology.org/E12-2021} {brat: a web-based tool for
  {NLP}-assisted text annotation}.
\newblock In \emph{Proceedings of the Demonstrations at the 13th Conference of
  the {E}uropean Chapter of the Association for Computational Linguistics},
  pages 102--107, Avignon, France. Association for Computational Linguistics.

\bibitem[{Stewart et~al.(2019)Stewart, Liu, and
  Cardell-Oliver}]{stewart:emnlp19}
Michael Stewart, Wei Liu, and Rachel Cardell-Oliver. 2019.
\newblock \href {https://doi.org/10.18653/v1/D19-3033} {{R}edcoat: A
  collaborative annotation tool for hierarchical entity typing}.
\newblock In \emph{Proceedings of the 2019 Conference on Empirical Methods in
  Natural Language Processing and the 9th International Joint Conference on
  Natural Language Processing (EMNLP-IJCNLP): System Demonstrations}, pages
  193--198, Hong Kong, China. Association for Computational Linguistics.

\bibitem[{Webson and Pavlick(2021)}]{albert:arxiv2021}
Albert Webson and Ellie Pavlick. 2021.
\newblock Do prompt-based models really understand the meaning of their
  prompts?
\newblock \emph{ArXiv}, abs/2109.01247.

\bibitem[{Wei et~al.(2013)Wei, Kao, and Lu}]{wei:nucleicacidsresearch13}
Chih-Hsuan Wei, Hung-Yu Kao, and Zhiyong Lu. 2013.
\newblock \href {https://doi.org/10.1093/nar/gkt441} {{PubTator: a web-based
  text mining tool for assisting biocuration}}.
\newblock \emph{Nucleic Acids Research}, 41(W1):W518--W522.

\bibitem[{Wei et~al.(2021)Wei, Bosma, Zhao, Guu, Yu, Lester, Du, Dai, and
  Le}]{wei2021flan}
Jason Wei, Maarten Bosma, Vincent~Y. Zhao, Kelvin Guu, Adams~Wei Yu, Brian
  Lester, Nan Du, Andrew~M. Dai, and Quoc~V. Le. 2021.
\newblock \href {http://arxiv.org/abs/2109.01652} {Finetuned language models
  are zero-shot learners}.
\newblock \emph{CoRR}, abs/2109.01652.

\bibitem[{Yang et~al.(2018)Yang, Zhang, Li, and Li}]{yang:acl18}
Jie Yang, Yue Zhang, Linwei Li, and Xingxuan Li. 2018.
\newblock \href {https://doi.org/10.18653/v1/P18-4006} {{YEDDA}: A lightweight
  collaborative text span annotation tool}.
\newblock In \emph{Proceedings of {ACL} 2018, System Demonstrations}, pages
  31--36, Melbourne, Australia. Association for Computational Linguistics.

\bibitem[{Yimam et~al.(2013)Yimam, Gurevych, Eckart~de Castilho, and
  Biemann}]{yimam:acl13}
Seid~Muhie Yimam, Iryna Gurevych, Richard Eckart~de Castilho, and Chris
  Biemann. 2013.
\newblock \href {https://aclanthology.org/P13-4001} {{W}eb{A}nno: A flexible,
  web-based and visually supported system for distributed annotations}.
\newblock In \emph{Proceedings of the 51st Annual Meeting of the Association
  for Computational Linguistics: System Demonstrations}, pages 1--6, Sofia,
  Bulgaria. Association for Computational Linguistics.

\bibitem[{Zhao et~al.(2021)Zhao, Wallace, Feng, Klein, and
  Singh}]{zhao:arxiv2021}
Tony~Z. Zhao, Eric Wallace, Shi Feng, Dan Klein, and Sameer Singh. 2021.
\newblock \href {http://arxiv.org/abs/2102.09690} {Calibrate before use:
  Improving few-shot performance of language models}.
\newblock \emph{CoRR}, abs/2102.09690.

\end{thebibliography}
\bibliographystyle{acl_natbib}

\appendix
\newpage
\clearpage

\section{Data and Statistics}

\begin{figure}
    \centering
    \includegraphics[width=0.5\textwidth]{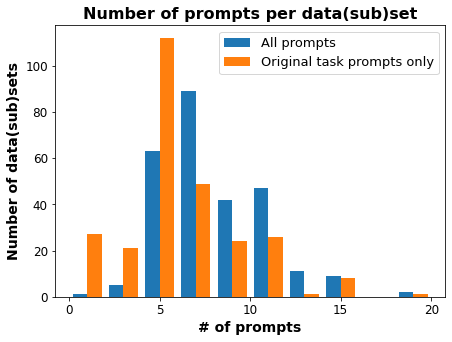}
    \caption{Most of the datasets have between 5 and 10 prompts.}
    \label{fig:nb_prompts}
\end{figure}

P3 is the largest public collection of English prompts and is actively growing. As of January 2022, it contains 2'052 English prompts for 170 English datasets (or 269 subsets, one dataset can contain multiple subsets with different prompts). There is an average of 7.6 prompts per data subset and an average 5.6 original-task prompts per data subset (see Figure~\ref{fig:nb_prompts}).

P3 was developed as part of the BigScience project for open research\footnote{\url{https://bigscience.huggingface.co}}. There was a open hackathon to collect prompts for as many English NLP dataset (or English subsets of datasets) as possible. Almost 50 unique contributors affiliated with more than 25 institutions in 10 countries participated. 

\section{Complete Views}

\begin{figure*}[t]
	\begin{center}
		\includegraphics[width=\linewidth]{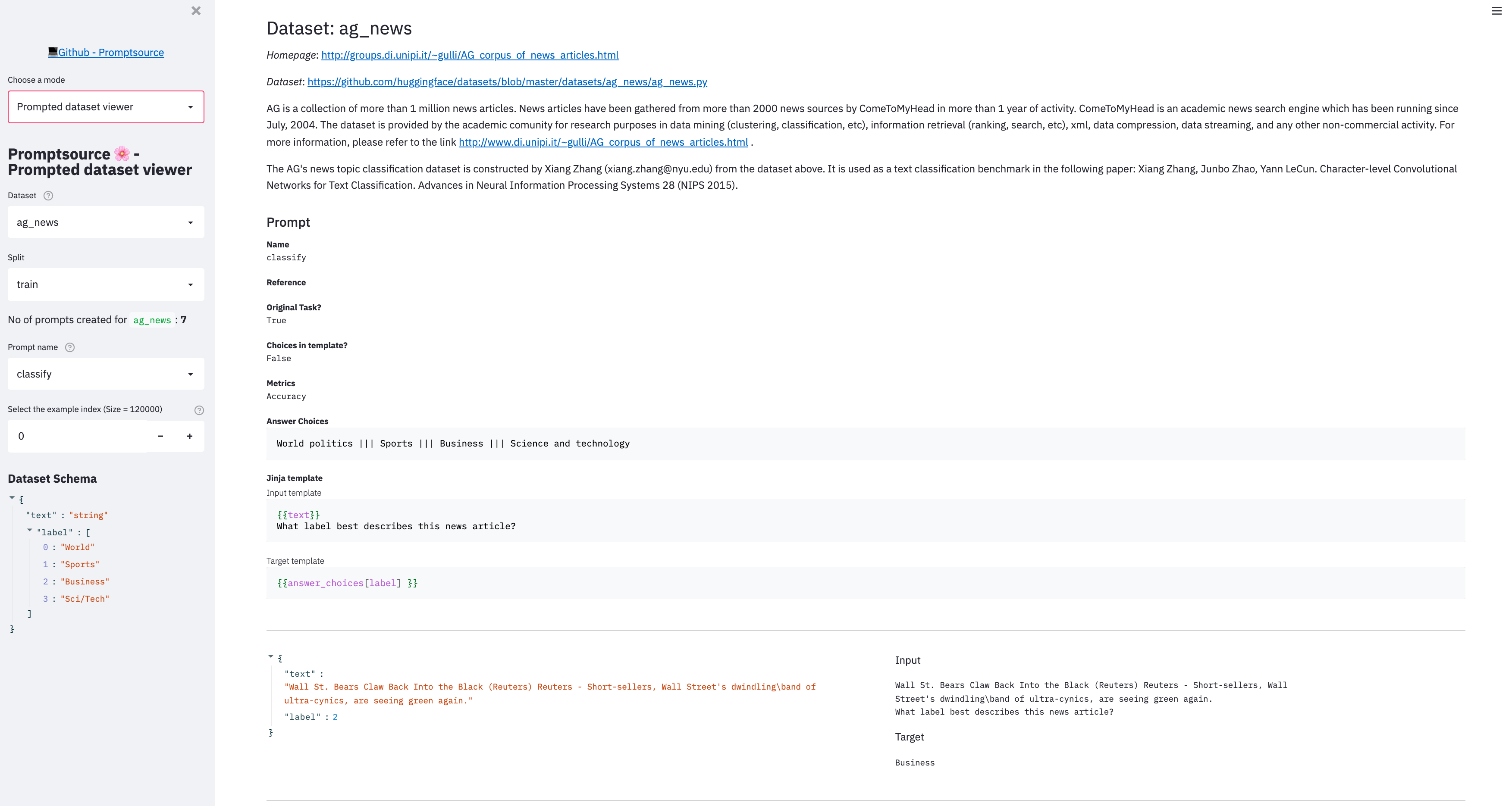}
	\end{center}
	\caption{Complete example of the \textit{Browse} view.}
	\label{fig:browse_entire}
\end{figure*}

\begin{figure*}[t]
	\begin{center}
		\includegraphics[width=\linewidth]{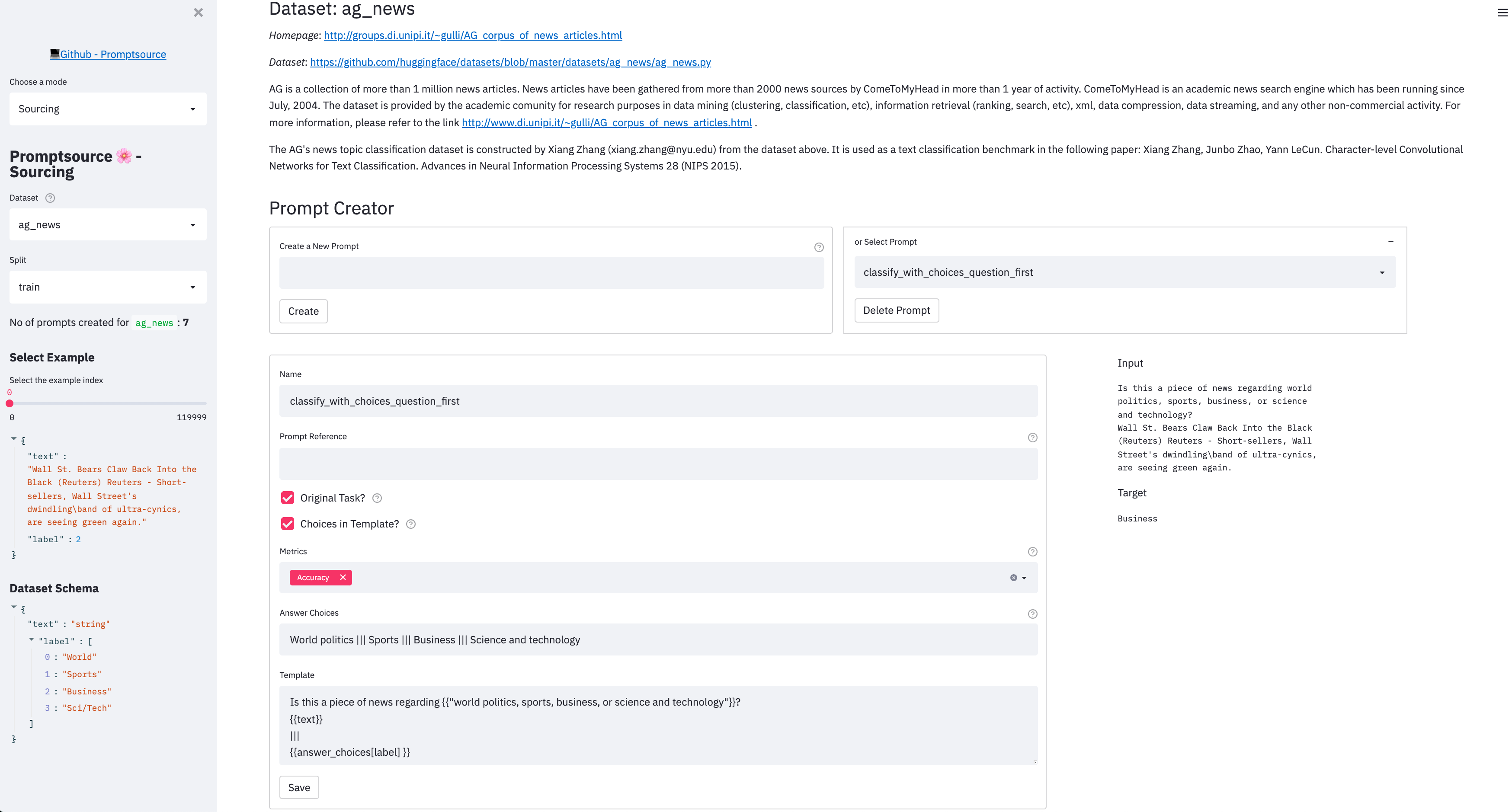}
	\end{center}
	\caption{Complete example of the \textit{Sourcing} view.}
	\label{fig:sourcing_entire}
\end{figure*}

\begin{figure*}[t]
    \begin{center}
        \includegraphics[width=\linewidth]{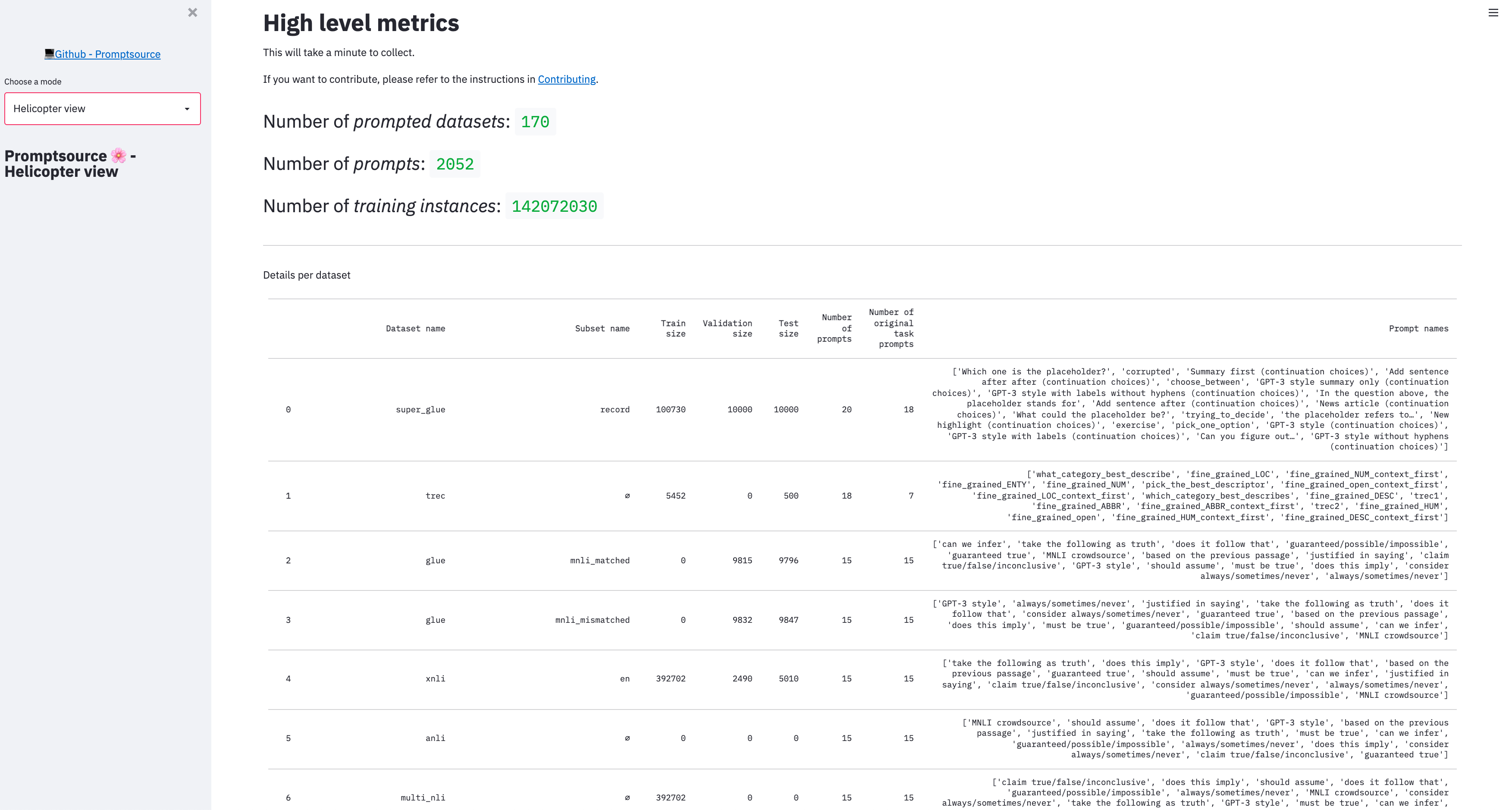}
    \end{center}
    \caption{Complete example of the \textit{Helicopter} view.}
    \label{fig:helicopter_entire}
\end{figure*}

We show higher resolution examples of the full interfaces for the \textit{Browse} (Figure~\ref{fig:browse_entire}), \textit{Sourcing} (Figure~\ref{fig:sourcing_entire}), and \textit{Helicopter} (Figure~\ref{fig:helicopter_entire}) views.

\end{document}